\documentclass{ecai} 
\usepackage{latexsym}
\usepackage{amssymb}
\usepackage{amsmath}
\usepackage{amsthm}
\usepackage{booktabs}
\usepackage{enumitem}
\usepackage{graphicx}
\usepackage{color}

\usepackage{multirow}
\usepackage{svg}
\usepackage{stfloats}

\definecolor{darkgreen}{RGB}{0, 172, 0}
\newcommand{\BibTeX}{B\kern-.05em{\sc i\kern-.025em b}\kern-.08em\TeX}

\begin{document}

%%%%%%%%%%%%%%%%%%%%%%%%%%%%%%%%%%%%%%%%%%%%%%%%%%%%%%%%%%%%%%%%%%%%%%%%

\begin{frontmatter}

%%% Use this command to specify your submission number.
%%% In doubleblind mode, it will be printed on the first page.

\paperid{9233} 

%%% Use this command to specify the title of your paper.

\title{Label Anything: Multi-Class Few-Shot \\ Semantic Segmentation with Visual Prompts}

%%% Use this combinations of commands to specify all authors of your 
%%% paper. Use \fnms{} and \snm{} to indicate everyone's first names 
%%% and surname. This will help the publisher with indexing the 
%%% proceedings. Please use a reasonable approximation in case your 
%%% name does not neatly split into "first names" and "surname".
%%% Specifying your ORCID digital identifier is optional. 
%%% Use the \thanks{} command to indicate one or more corresponding 
%%% authors and their email address(es). If so desired, you can specify
%%% author contributions using the \footnote{} command.

\author[A]{\fnms{Pasquale}~\snm{De Marinis}\orcid{0000-0001-8935-9156}\thanks{Corresponding Author. Email: pasquale.demarinis@uniba.it}}
\author[A]{\fnms{Nicola}~\snm{Fanelli}\orcid{0009-0007-6602-7504}}
\author[A]{\fnms{Raffaele}~\snm{Scaringi}\orcid{0000-0001-7512-7661}}
\author[A]{\fnms{Emanuele}~\snm{Colonna}\orcid{0009-0009-0932-3424}}
\author[B]{\fnms{Giuseppe}~\snm{Fiameni}\orcid{0000-0001-8687-6609}}
\author[A]{\fnms{Gennaro}~\snm{Vessio}\orcid{0000-0002-0883-2691}}
\author[A]{\fnms{Giovanna}~\snm{Castellano}\orcid{0000-0002-6489-8628}} 

\address[A]{Department of Computer Science, University of Bari Aldo Moro, Bari, Italy}
\address[B]{NVIDIA AI Technology Center, Bologna, Italy}

%%% Use this environment to include an abstract of your paper.

\begin{abstract}
Few-shot semantic segmentation aims to segment objects from previously unseen classes using only a limited number of labeled examples. In this paper, we introduce Label Anything, a novel transformer-based architecture designed for multi-prompt, multi-way few-shot semantic segmentation. Our approach leverages diverse visual prompts---points, bounding boxes, and masks---to create a highly flexible and generalizable framework that significantly reduces annotation burden while maintaining high accuracy.
Label Anything makes three key contributions: (\textit{i}) we introduce a new task formulation that relaxes conventional few-shot segmentation constraints by supporting various types of prompts, multi-class classification, and enabling multiple prompts within a single image; (\textit{ii}) we propose a novel architecture based on transformers and attention mechanisms; and (\textit{iii}) we design a versatile training procedure allowing our model to operate seamlessly across different $N$-way $K$-shot and prompt-type configurations with a single trained model.
Our extensive experimental evaluation on the widely used COCO-$20^i$ benchmark demonstrates that Label Anything achieves state-of-the-art performance among existing multi-way few-shot segmentation methods, while significantly outperforming leading single-class models when evaluated in multi-class settings. Code and trained models are available at \url{https://github.com/pasqualedem/LabelAnything}.
\end{abstract}

\end{frontmatter}

%%%%%%%%%%%%%%%%%%%%%%%%%%%%%%%%%%%%%%%%%%%%%%%%%%%%%%%%%%%%%%%%%%%%%%%%

\section{Introduction}
\label{sec:intro}

\begin{figure*}[t]
    \centering
    \includegraphics[width=\linewidth]{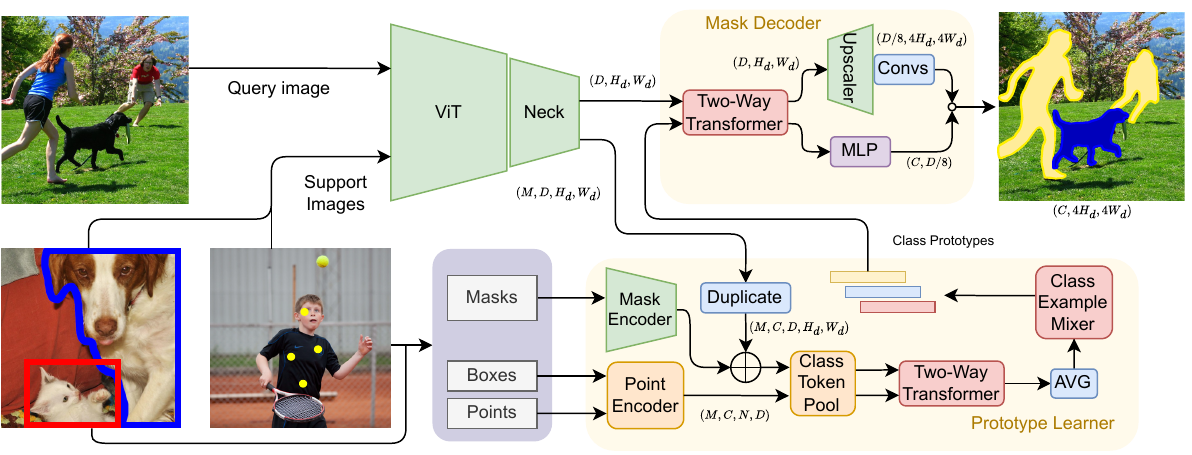}
    \caption{Overview of the proposed framework, illustrating the workflow for a scenario involving three classes, two support images, and all three types of prompts. Label Anything leverages a pre-trained ViT to derive image features. Prompts are fed into the Prototype Learner, which generates a distinct prototype for each class. Subsequently, these prototypes inform the Mask Decoder, facilitating accurate pixel classification across the query image.}
    \label{fig:LA}
\end{figure*}

The progress of semantic segmentation has been closely tied to the availability of extensive labeled datasets, which often pose a significant challenge in deploying these systems for real-world applications. Typically, models trained on such datasets excel at recognizing and classifying objects they were trained on but falter when encountering new concepts or domains, thus limiting their practical utility. Segmenting objects from previously unseen classes with few or no labeled examples remains a pivotal challenge in computer vision.

Few-shot semantic segmentation (FSS) has gained attention as the task of segmenting a query image based on a limited set of examples \cite{shaban2017one}. While the majority of research in FSS has concentrated on binary segmentation tasks (distinguishing background from foreground), there has been limited exploration of methods capable of segmenting multiple classes simultaneously, a step toward more flexible and \textit{domain-agnostic} segmentation \cite{dong2018few,liu2020dynamic,liu2020part,tian2020differentiable,wang2019panet,zhang2022mfnet}. Furthermore, aside from an initial exploration by Rakelly et al.~\cite{Rakelly2018ConditionalNF}, which considered coordinates as prompts, the field has predominantly focused on mask annotations as the sole input type. This underscores the growing need to support diverse input annotations, reducing reliance on extensive human annotation and facilitating smoother integration with other models, such as object detectors that generate bounding boxes.

This work presents Label Anything (LA), a novel deep neural network architecture designed for multi-prompt, multi-class FSS. Our method adopts a \textit{prototype}-based learning paradigm, utilizing diverse visual prompts---points, bounding boxes, and masks---to establish a highly flexible and generalizable framework (Fig.~\ref{fig:LA} illustrates our approach). This innovation not only addresses the challenge of segmenting objects from unseen classes with limited examples but also significantly reduces the annotation burden. Building upon Segment Anything's use of diverse prompts for segmentation \cite{kirillov2023segment}, our method significantly enriches FSS by enhancing the adaptability of the support set and extending the framework to multi-class challenges.

Unlike traditional approaches that focus on binary segmentation and adhere to strict class example limits, LA supports various prompt types and a flexible number of class examples per support set. This flexibility enables the capture of complex inter-class relationships from a single image, thereby improving segmentation precision. Moreover, LA's architecture is designed to learn from different support set configurations and prompt types within a single training cycle, thereby eliminating the need for retraining across various FSS scenarios. Such versatility ensures that LA can be applied across a wide range of domains, offering a comprehensive solution that moves beyond the constraints of conventional FSS models.

Our extensive experimental analysis on the widely used COCO-$20^i$ benchmark \cite{nguyen2019feature,wang2019panet} shows that our model achieves state-of-the-art performance among existing multi-way FSS methods. Moreover, it significantly outperforms leading single-class FSS models when evaluated in multi-class settings, highlighting their limited adaptability to real-world scenarios where multiple object categories must be segmented simultaneously. These results underscore the need for dedicated multi-class FSS solutions. Beyond its technical contributions, LA offers practical advantages by enabling more accessible and efficient semantic segmentation, particularly in contexts where annotated data is scarce.

In summary, this work makes three contributions to advancing the field of semantic segmentation:
\begin{itemize}
    \item We introduce a new task, multi-prompt multi-way few-shot semantic segmentation, which emerges from a substantial relaxation of conventional few-shot segmentation constraints. This task enhances the applicability of segmentation models by supporting various types of prompts and multi-class classification, and enables multiple prompts within a single image, thereby overcoming the limitations of prior methods that require one image per prompt. 
    \item We propose a novel architecture centered on attention mechanisms and transformer blocks, which assume the primary computational role. This design departs from traditional convolution-centric approaches, enabling more flexible and scalable solutions for segmentation tasks.
    \item We design a versatile training procedure that enables our model to be trained once and operate seamlessly across various $N$-way $K$-shot and prompt-type configurations, unlike prior methods that require separate models for each parameter combination. 
\end{itemize}
% Together, these contributions mark a significant step toward adaptable and practical few-shot segmentation models suited for real-world applications.

The rest of this paper is organized as follows. Section~\ref{sec:related_work} reviews related work. Section \ref{sec:preliminary} introduces the task of multi-prompt multi-way FSS derived from a relaxation of the typical FSS constraints. Section~\ref{sec:label_anything} details our proposed approach. Section~\ref{sec:experiments} evaluates our model's performance. Section~\ref{sec:conclusion} concludes the paper with a summary of our findings and directions for future research.

\section{Related Work}
\label{sec:related_work}

\paragraph{Few-Shot Semantic Segmentation}
Semantic segmentation, which involves assigning a class label to each pixel in an image, is a fundamental task in computer vision. The introduction of convolutional neural networks (CNNs), with pioneering architectures such as the Fully Convolutional Network \cite{long2015fully} and U-Net \cite{ronneberger2015u}, was pivotal in advancing the field. These architectural innovations laid the foundation for numerous subsequent methods, each tailored to address specific challenges (e.g., \cite{cciccek20163d,oktay2018attention}).

Building upon few-shot learning principles \cite{finn_model-agnostic_2017,snell_prototypical_2017,sung_learning_2018,vinyals_matching_2017}, few-shot semantic segmentation has evolved as a significant extension of semantic segmentation, achieving high accuracy with minimal reliance on large, annotated datasets. FSS involves per-pixel classification in a query image, guided by a ``support set'' typically comprising a small number of images and their corresponding mask annotations. This task hinges on the model's ability to leverage support set information for accurate segmentation.

Early attempts in FSS utilized CNNs to synthesize support set information. This information was integrated into the segmentation process by modifying weights and biases or merging it with query image features \cite{Rakelly2018ConditionalNF,shaban2017one}. This foundational work laid the groundwork for more sophisticated approaches broadly categorized into \textit{prototype}-based and \textit{affinity}-based learning. Prototype-based methods focus on deriving one or several representative prototypes for the concept to be segmented, applying these to the query image through distance metrics or decoder networks \cite{dong2018few,li2021adaptive,liu2020part,liu2022intermediate,wang2019panet,zhang2019canet,zhang2022mfnet,zhang2020sg}. Most existing approaches employ CNNs as feature extractors; more recently, FPTrans \cite{zhang2022feature} explored the use of Vision Transformers (ViT) \cite{dosovitskiy2020image} as encoder. In contrast, affinity learning approaches delve into the pixel-wise relationships between support and query features \cite{chen2024pixel,min2021hypercorrelation,shi2022dense,wang2023rethinking,zhang2021few}. While affinity-based approaches often achieve stronger performance, they come at a high computational cost. These methods require access to and processing of the entire support set during inference for each query image, making them challenging to scale---especially in multi-class (multi-way) settings. In fact, all existing multi-way methods are prototype-based, as they enable precomputed, reusable class representations, making them more scalable and efficient in real-world scenarios.

% In recent years, considerable interest has been shown in FSS. 
However, most recent developments have focused on a binary paradigm, where images are segmented based on a single concept specified by the support set. The binary (single-way) and multi-class FSS settings present distinct challenges. Binary classification models struggle to adapt to multi-class scenarios without a significant performance drop, highlighting the need for specialized solutions. As a result, only a few studies have explored the complex area of multi-class FSS \cite{dong2018few,liu2020dynamic,liu2020part,tian2020differentiable,wang2019panet,zhang2022mfnet}. This involves enabling the support set to introduce multiple classes for segmentation, requiring the model to distinguish between $N$ foreground classes in the query image, leveraging a set of $K$ support images per class, solving a prototype-based $N$-way $K$-shot task.
% a departure from the more straightforward foreground-background dichotomy. Multi-class FSS typically adopts an $N$-way $K$-shot framework, entailing $N$ classes within the support set and $K$ instances per class. This approach, designed for segmenting multiple classes, relies on a prototype-based model architecture to encapsulate and communicate class-specific information throughout the network effectively.
For example, MFNet \cite{zhang2022mfnet} leverages masked global pooling alongside attention mechanisms to generate class prototypes that integrate with query image features for segmentation. To our knowledge, MFNet represents the most recent effort explicitly addressing multi-way FSS---a direction that has since been largely overlooked in the literature. Despite this, the multi-class formulation is arguably more practical and relevant to real-world applications than single-class segmentation, which considers only one object category at a time.

Nevertheless, existing methods suffer from two main limitations: (\textit{i}) the performance of multi-way approaches critically depends on the predefined values of $N$ and $K$, which must be fixed prior to training---this also impacts many single-way models by constraining the number of examples per segmented concept; (\textit{ii}) these methods typically treat different classes within the same image independently, repeating the same image for each class, resulting in a more complex formulation and increased computational overhead. To address these limitations, we revisit the multi-class FSS task from a more flexible and scalable perspective. By relaxing rigid assumptions and architectural constraints, we propose a more adaptable solution that can handle diverse and flexible segmentation scenarios.

\paragraph{Promptable Segmentation}
Integrating \textit{prompt}-based learning, inspired by its success in natural language processing, into computer vision introduced a novel paradigm. In particular, a significant breakthrough came from the Segment Anything Model (SAM), introduced by Kirillov et al.~\cite{kirillov2023segment}. 
% This innovation harnesses an interactive, prompt-guided vision framework, representing a pivotal moment reminiscent of the GPT series but in computer vision. SAM has been trained on a vast dataset comprising over 1 billion masks extracted from 11 million images, employing a ``promptable'' segmentation task. This innovative approach equips SAM with the unique capability for robust zero-shot generalization, enabling it to excel in various tasks such as edge detection and instance segmentation.
% In recent times, 
The research community has been prolific in extending the capabilities of SAM and exploring its potential across diverse domains. These endeavors have pushed the boundaries of what SAM can accomplish and have successfully applied it to a wide range of tasks, including medical image analysis \cite{mazurowski2023segment}, image inpainting \cite{yu2023inpaint}, and image editing \cite{xie2023edit}. Recent innovations, such as SegGPT \cite{wang2023seggpt} and SEEM \cite{zou2024segment}, have extended the use of prompts to guide predictions, exploring the integration of textual and audio prompts to enhance model performance across diverse tasks.

However, despite these advancements, a gap remains in applying these models to FSS, particularly in segmenting objects into previously unseen classes based on a limited number of annotated examples. While models like SegGPT show promise in FSS scenarios, their generalization capabilities are often attributed to the vast datasets used for training, rather than an inherent ability to leverage support set information effectively. Furthermore, apart from an early exploration by Rakelly et al.~\cite{Rakelly2018ConditionalNF}, research in FSS has scarcely investigated the use of diverse annotations within support sets---such as points or bounding boxes---as alternatives to traditional mask prompts, leaving a substantial area of FSS largely unexplored. For this reason, we relax the usage of the binary mask only as support information by introducing bounding boxes and points. Specifically, we design a new prototype learning stage, in which diverse prompts are encoded into a common, shared representation, thereby enhancing our model's flexibility to accommodate various types of prompts. 
% To address this, we relax the restriction to binary masks by incorporating bounding boxes and point annotations as alternative supervision signals. Specifically, we propose a new prototype learning stage in which diverse prompt types are encoded into a unified representation space, thereby enhancing the model's flexibility and robustness to heterogeneous support annotations.

\section{Multi-Prompt Multi-Way FSS}
\label{sec:preliminary}

We introduce the task of \textit{multi-prompt}, \textit{multi-way}, few-shot semantic segmentation. While the standard single-prompt, multi-way few-shot task involves classifying each pixel in a query image into one of $N$ foreground classes using $N \times K$ support image-mask pairs---with one mask per image---our approach extends this setting. Specifically, we classify the pixels of a query image $I_q$ into $N$ classes using $L$ examples, allowing for a variable number and type of shots per class, as well as the inclusion of multiple classes within a single support image, which is not possible in the traditional setting. Furthermore, we extend beyond the sole usage of masks as a prompt modality by introducing multiple prompt types, which can be either points or bounding boxes, to reduce human effort in shot labeling significantly.
% To support this flexible setting, we adopt episodic training \cite{vinyals_matching_2017}, where each training episode consists of a query image and a corresponding support set.}
% We introduce the task of \textit{multi-prompt}, \textit{multi-way}, few-shot semantic segmentation. \raff{Despite the well known \textit{single-prompt} \textit{multi-way} few shot task, in which each pixel of a query image is discriminated over $N$ foreground classes, using $N \times K$ support images-masks, where a single image is an example for a given class, our novel approach} involves classifying the pixels of a query image $I_q$ into one of $N$ classes, using $L$ examples, allowing different number and type of shots per class and multiple classes in a single image. To achieve this goal, we employ "episodic training" \cite{vinyals_matching_2017}, which involves training episodes that consist of a query image and a distinct support set.

To tackle few-shot tasks, the dataset $\mathcal{D}$ must be partitioned into two subsets: a training set $\mathcal{D}_{\text{train}}$ and a test set $\mathcal{D}_{\text{test}}$. Crucially, we define two distinct sets of segmentation categories, $\mathcal{C}_{\text{seen}}$ for training and $\mathcal{C}_{\text{unseen}}$ for testing, ensuring $\mathcal{C}_{\text{seen}} \cap \mathcal{C}_{\text{unseen}} = \emptyset$. Annotations in $\mathcal{D}_{\text{train}}$ utilize classes from $\mathcal{C}_{\text{seen}}$, whereas $\mathcal{D}_{\text{test}}$ is evaluated on classes from $\mathcal{C}_{\text{unseen}}$, representing concepts not encountered during training to assess the model's generalization capabilities in a few-shot setting.

During each training ``episode'', the model receives a query image and its corresponding ground truth pair $(I_q, M_q)$, where $I_q \in \mathbb{R}^{3\times H\times W}$ represents the RGB query image and $M_q \in \mathbb{R}^{(N+1)\times H\times W}$ is the segmentation mask. The mask $M_q$ assigns each pixel in $I_q$ to one of $N+1$ classes, including the background. In addition, the model is provided with a support set $\mathcal{S} = \{(I_i, A_i)\}_{i=1}^L$, consisting of $L$ image-prompt annotation pairs that define the concepts to be segmented. Each pair $(I_i, A_i)$ is referred to as a ``shot'', with $L$ typically being a small number. In multi-class FSS tasks, $L$ is commonly set as $N \times K$, where $N$ is the number of classes and $K$ is the number of examples (shots) per class. 
% Standard evaluation protocols on benchmark datasets usually adopt values for $N$ and $K$ from the set $\{1, 2, 5\}$.

In more detail, we have $A_i = \{(M_{i,n}, P_{i,n}, B_{i,n})\}_{n=1}^N$, where $M_{i,n}$ is a binary mask for class $n$, $P_{i,n}$ represents point coordinates, and $B_{i,n}$ denotes bounding boxes, harnessing the multi-prompt setting. 
% \textbf{The prompt type for the $n$-th class in image $I_i$ is randomly selected during training, with padding applied as necessary for absent classes or prompt types. (probabilmente da rimuovere da qui)} Evaluation on $\mathcal{D}_{\text{test}}$ critically involves annotations from the support set in $\mathcal{C}_{\text{unseen}}$, which tests the model's adaptability to new classes.
Our approach significantly extends the flexibility of FSS by allowing for multiple types of prompts within the support set. Unlike previous FSS methodologies, LA does not require the support set size $L$ to match $N \times K$ strictly, enabling variable numbers of shots per class. Additionally, LA permits a single image to serve as a support example for multiple classes without repetition, linking different class prompts to the same image, thus maximizing the exploitation of inter-class relationships within annotations, enhancing flexibility and reducing human effort in shot labeling.

\section{Proposed Method}
\label{sec:label_anything}

% We introduce \textit{Label Anything} (LA), a model capable of labeling any image across any class using a minimal set of prompts for each class. This approach does not prescribe a predefined number of prompts per class; instead, it mandates that the support set comprehensively covers all required classes.

We introduce \textit{Label Anything}, a model for few-shot segmentation tailored to the multi-prompt, multi-way setting. Given a support set consisting of a few images with visual prompts for each class, LA enables the semantic segmentation of user-specified target classes. The overall framework is illustrated in Fig.~\ref{fig:LA}. This section details the model architecture, training procedure, and key design choices that allow LA to generalize effectively to diverse and previously unseen classes. For clarity, we omit the batch size dimension (corresponding to the episode axis) in the following explanation; it is reintroduced when describing the training procedure in Sec.~\ref{sec:training_procedure}.

% \raff{\textbf{I would move this sentence before the previous one for clarity.}}
% In more detail, we have $A_i = \{(M_{i,n}, P_{i,n}, B_{i,n})\}_{n=1}^N$ for each support set image, where $M_{i,n}$ is a binary mask for class $n$ in image $i$, $P_{i,n}$ represents point coordinates, and $B_{i,n}$ denotes bounding boxes. The prompt type for class $n$ in image $i$ is randomly selected during training, with padding applied as necessary for absent classes or prompt types. Evaluation on $\mathcal{D}_{\text{test}}$ involves support set annotations from $\mathcal{C}_{\text{unseen}}$, testing the model's adaptability to new classes.

% \begin{figure*}[!ht]
%     \centering
%     \caption{Class prototype generation process. LA begins with extracting image features from a single support set example using ViT. This example might include multiple prompts related to different classes and types. A CNN-based Mask Encoder processes dense prompts (masks), while point coordinates and bounding boxes are transformed into embeddings via the Point Encoder, with bounding boxes treated as pairs of point embeddings. These processed prompts are then integrated with the image features. The Class Token Pool enriches dense and sparse embeddings with class information to infuse class-specific semantics. The subsequent two-way cross-attention between dense and sparse features yields class prototypes for each example.}
%     \resizebox{0.95\textwidth}{!}{\includegraphics[width=\linewidth]{figures/PL.pdf}}\label{fig:prototype_generation}
% \end{figure*}

\begin{figure}[!ht]
    \centering
    \includegraphics[width=\linewidth]{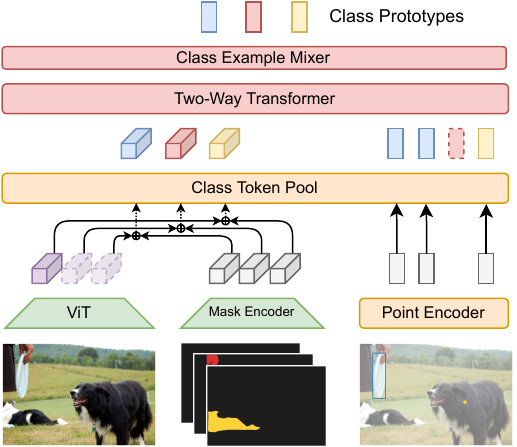}\label{fig:prototype_generation}
    % \caption{Class prototype generation process. LA begins with extracting image features from a single support set example using ViT. This example might include multiple prompts related to different classes and types. A CNN-based mask encoder processes dense prompts (masks), while point coordinates and bounding boxes are transformed into embeddings via the point encoder, with bounding boxes treated as pairs of point embeddings. These processed prompts are then integrated with the image features. The class token pool enriches dense and sparse embeddings with class information to infuse class-specific semantics. The subsequent two-way cross-attention between dense and sparse features yields class prototypes for each example.}
    \caption{
    Class prototype generation. LA extracts image features from a single support example via ViT. Dense prompts (masks) are encoded with a CNN; point coordinates and boxes are embedded via a point encoder, with boxes as point pairs. Prompt embeddings are fused with image features. A class token pool enriches them with class semantics. Two-way cross-attention between dense and sparse features produces the class prototypes. Class Example Mixer allows class information exchange.
    }
\end{figure}

% Label Anything processes a query image $I_q$ for segmentation alongside a support set $\mathcal{S}$, utilizing three core components in its architecture, the Image Encoder, the Prompt Encoder, and the Mask Decoder, as illustrated in Fig.~\ref{fig:LA}.

\subsection{Feature Extraction and Neck}

We extract features from both the query image and the support set using a Vision Transformer, specifically ViT-B/16~\cite{dosovitskiy2020image}, initialized with ImageNet pretraining in line with common practice in FSS~\cite{zhang2022feature}. By employing a ViT, we avoid aggregating multi-level feature maps from different layers, as is typically required with CNN-based encoders. Instead, we use the output from the final layer of the ViT, resulting in feature maps denoted as $F_q$ for the query and $F_{\mathcal{S}}$ for the $L$ support images. These have dimensionality $D_{\text{ViT}} \times H_d \times W_d$, where $H_d = H / x_p$ and $W_d = W / x_p$, with $x_p$ representing the patch size (e.g., $16$ for ViT-B/16).

To reduce the channel dimensionality and facilitate subsequent processing, we append a convolutional neck to the ViT. This module reduces the feature dimension from $D_{\text{ViT}} = 768$ to a more compact representation with $D = 256$ channels, yielding the final image features used in the following stages of the model.

\subsection{Prototype Learning}

Inspired by the architecture of the Segment Anything Model~\cite{kirillov2023segment}, we design our model to handle multiple types of prompts---masks, bounding boxes, and points---applied to the support images to define the concepts to be segmented in the query image. To this end, we extend the prompt encoder mechanism from SAM to suit the requirements of few-shot semantic segmentation.

\paragraph{Prompt Encoding}
We begin by extracting embeddings from the input prompts. Mask prompts are processed through a series of convolutional layers, producing dense embeddings denoted as $P_d$. Point and box prompts are handled by a point encoder, which applies positional encodings to the $(x, y)$ coordinates of the input points, resulting in sparse embeddings $P_s$. We introduce three learnable type-specific encodings corresponding to point prompts, box top-left corners, and box bottom-right corners to distinguish between different types of sparse prompts. These are added to the respective sparse embeddings. All prompt embeddings---dense and sparse---are projected to have a dimensionality of $D$, matching the channel dimension of the image feature maps, which facilitates subsequent integration within the network. Once these embeddings are obtained, we contextualize the sparse prompts by allowing them to exchange information via a self-attention layer followed by a residual connection. This attention mechanism operates independently on the prompts of each support image, without sharing information across different support examples.

A key distinction from SAM lies in the fact that, in our setting, prompts may correspond to different semantic classes that the user wishes to segment. To account for this, we introduce a \textit{token pool}, implemented as a learnable Gaussian-initialized matrix of size $R \times D$, where $R$ is a large integer greater than the maximum number of classes $N$ used during testing. For each episode (i.e., a support-query pair), one row from this matrix is sampled per class and added to the corresponding dense and sparse embeddings. This mechanism provides the model with explicit information about which prompts belong to the same class, thereby improving class-level disentanglement and representation consistency during segmentation.

\paragraph{Image-Prompt Fusion}
After extracting representations for the support images and their corresponding prompts, we aim to fuse this information to construct \textit{class prototypes} that can effectively guide the segmentation of the query image.

To this end, we adopt a two-way transformer mechanism inspired by SAM, where multi-head attention is applied bidirectionally, from the sparse prompts to the image features and vice versa. Information exchange is restricted to each prompt and its associated support image. Before this fusion step, we incorporate the dense prompts (i.e., masks) into the image features via element-wise addition, yielding enriched image representations defined as
\begin{equation}
F_{l,c} = F_l + P_d^{l,c}, \quad \text{for } l = 1, \dots, L,
\end{equation}
where each class $c$ is associated with a distinct feature map.

The fusion process is then carried out using the following multi-head attention layers:
\begin{align}
P_s^{l,c} &= \text{MultiHead}(P_s^{l,c}, F_{l,c} + PE, F_{l,c} + PE), \\
F_{l,c} &= \text{MultiHead}(F_{l,c} + PE, P_s^{l,c}, P_s^{l,c}),
\end{align}
where $\text{MultiHead}(Q, K, V)$ denotes the standard multi-head attention operator with queries $Q$, keys $K$, and values $V$. Here, $P_s^{l,c}$ represents the sparse prompt embeddings for class $c$, and $PE$ denotes positional encodings adapted to the spatial dimensions $(H_d, W_d)$.

After fusion, we retain only the updated dense features $F_{l,c}$, which now encapsulate information from both the support image and the sparse prompts. We apply global average pooling over the spatial dimensions to obtain a compact representation, producing a class-example embedding $e_{l,c} \in \mathbb{R}^D$ for each support image and target class. Before aggregating these embeddings across support examples, we introduce the \textit{class-example mixer}, a self-attention block designed to refine class representations. For each class $c$, we concatenate the embeddings of its support examples into a matrix $E_c = \left[\begin{smallmatrix} e_{1,c} & \dots & e_{L,c} \end{smallmatrix}\right]$, and apply the following operation:
\begin{equation}
e_c = \text{MeanPool}(\text{SelfAttn}(E_c)),
\end{equation}
where $\text{SelfAttn}$ denotes a self-attention layer applied across the example dimension. The resulting output $e_c$ serves as the class prototype, capturing a rich, aggregated representation of the class to be segmented. These prototypes are subsequently used during the query image mask decoding process.

\subsection{Mask Decoding}
\label{sec:mask_decoder}

The mask decoder takes as input the class prototypes $E = [e_c]_{c=1}^N$ and the query image features $F_q$, and produces a segmentation mask for each target class.
This process begins with a two-way transformer module, which serves a different purpose than in the prototype learner. Its goal is to facilitate pattern matching between the class prototypes $E$ and the pixel locations within the query image $F_q$ by enabling mutual attention. This reciprocal interaction transfers relevant class-specific information to the query features, guiding segmentation. Formally, this attention mechanism is defined as:
\begin{align}
F_q &= \text{MultiHead}(F_q, E, E),\\
E &= \text{MultiHead}(E, F_q, F_q).
\end{align}

Next, the query features are upsampled to a spatial resolution of $(4H_d, 4W_d)$, and their channel dimension is reduced to $D/8$ using a sequence of transposed convolutional layers. In parallel, the class prototypes are projected to the same feature dimension using an MLP, ensuring alignment with the transformed query features. The refined query features $F_q$ are then further processed through three spatial convolutional layers with kernel size $3 \times 3$, which enhance the spatial coherence and smoothness of the segmentation output.

As a result, we obtain the final query representation $F_q \in \mathbb{R}^{{4H_d} \times 4W_d \times \frac{D}{8}}$ and the aligned class prototype matrix $E \in \mathbb{R}^{N \times \frac{D}{8}}$. The predicted segmentation mask is computed as follows:
\begin{equation}
\hat{M} = F_q \cdot E^{\top},
\end{equation}
where each element $\hat{M}_{xyn}$ represents the logit for pixel $(x, y)$ to belong to the $n$-th class.

\subsection{Training Procedure}
\label{sec:training_procedure}

One of the primary objectives of this work is to enable our model to be trained \textit{once} for applications across arbitrary combinations of $N$ and $K$ during testing, as well as for various prompt inputs. To accomplish this, we introduce an innovative episodic training approach that optimizes resource utilization while accommodating diverse $N$, $K$, and prompt-type configurations.

The procedure leverages ``episodic training'' \cite{vinyals_matching_2017}, having flexible number of classes and shots, reflecting a real world scenario. Specifically, we defined potential batch configurations as $(B, N, K)$, where $B$ represents the batch size, indicating the number of learning episodes per batch. The selection of $B$ is tailored to the $N$ and $K$ parameters to ensure optimal GPU resource management, utilizing powers of 2. Specifically, $B \times N \times K + B$ represents the aggregate image count per batch, encompassing both support and query images. Each training batch is then constructed by randomly selecting a $(B, N, K)$ tuple.

For each episode in a training batch, $N$ classes are randomly sampled from the set of seen classes, $\mathcal{C}_{\text{seen}}$. A query image is then selected such that, with uniform probability, it contains either a subset or all of the $N$ chosen classes. If no query image satisfies the desired class configuration, $N$ is adjusted accordingly. This strategy enables the construction of episodes where the query image does not necessarily include all the classes in the support set, thereby mitigating a well-known bias in FSS research \cite{kang2022integrative}. Subsequently, $K$ support examples are sampled per class. We identify all annotations corresponding to the selected $N$ classes within each chosen image, allowing multiple annotations per image. This flexibility ensures our model's readiness for any $L$ image-annotation pair configuration, where $L$ may deviate from the strict $N \times K$ formula.

To increase variability during training, each episode includes a random selection of prompt types for the instance annotations in the support set. For instances requiring point-based annotations, points are sampled in proportion to the area of the instance mask. This approach tailors supervision to the specific characteristics of each episode and prompt type, leading to more robust model learning.

\section{Experimental Evaluation}
\label{sec:experiments}

We evaluated LA across two distinct FSS tasks: $1$-way $1$-shot segmentation and $N$-way $K$-shot segmentation. Performance is reported as mean Intersection-over-Union (mIoU) across test classes.

\subsection{Dataset and Validation}

Our assessment was based on the COCO-$20^i$ dataset \cite{nguyen2019feature,wang2019panet}, a prominent FSS benchmark sourced from the MS COCO dataset \cite{lin2014microsoft}, to evaluate our model across both $1$-way $1$-shot and $N$-way $K$-shot segmentation tasks. COCO-$20^i$ amplifies the challenge by dividing its 80 classes into four distinct folds, each encompassing 20 classes. Employing cross-validation, we rotated classes within each fold as $\mathcal{C}_{\text{unseen}}$, contrasting with $\mathcal{C}_{\text{seen}}$ for the remaining classes. This setup aligned $\mathcal{D}_{\text{train}}$ and $\mathcal{D}_{\text{test}}$ with the original dataset's arrangement, ensuring our model faced novel images and classes during testing. Our model was fine-tuned in a single iteration per cross-validation phase, demonstrating the effectiveness of our training strategy.

To ensure a rigorous comparison with established benchmarks, especially in binary FSS contexts \cite{min2021hypercorrelation,nguyen2019feature}, we adopted a consistent evaluation methodology. We conducted tests on 1,000 randomly chosen episodes, using five distinct random seeds for each and averaging the results. This procedure was replicated for $N$-way $K$-shot evaluations, aligning with methodologies from prominent studies \cite{liu2020part,wang2019panet,zhang2022mfnet}. Unless otherwise stated, we limited our prompt type to masks during tests for more accurate benchmarking, facilitating direct comparisons across different approaches and configurations.

\subsection{Setting}

% We adopted a uniform input resolution of $1024 \times 1024$ for query and support images, resizing images and applying padding to maintain the aspect ratio. Following the methodology proposed by Zhang et al.~\cite{zhang2022feature}, we used the ViT-B/16 model \cite{dosovitskiy2020image}, augmented with Segment Anything \cite{kirillov2023segment} pretraining and with its weights kept frozen, serving as our backbone. This configuration yielded a singular feature map per image. Class-specific masks were supplied at a refined resolution of 256 $\times$ 256, while points and bounding boxes were integrated into the model using normalized coordinates.

We adopted a uniform input resolution for query and support images by resizing them to $480 \times 480$. We used the ViT-B/16 architecture as a backbone, pre-trained with the masked autoencoder (MAE) strategy \cite{cheng2022masked} on ImageNet. 
Our training strategy optimized the focal loss \cite{lin2017focal}, which mitigates class imbalance by emphasizing hard-to-classify examples. To further account for intra-image class imbalance, we applied per-class weighting proportional to the number of pixels belonging to each class in the query image.
% This pretraining approach preserves spatial information, making it well-suited for segmentation tasks.
% Our training strategy optimized the focal loss \cite{lin2017focal}, which addresses class imbalance by focusing on difficult-to-classify examples.
% computed for a single training instance as:
% \[
% \mathcal{L} = \frac{1}{N}\sum_n^{N}{\left[w_n \cdot (1 - e^{-l_{ce}(\hat{y}_n, y_n)})^{\gamma} \cdot l_{ce}(\hat{y}_n, y_n)\right]},
% \]
% where $N$ is the number of classes, $w_n$ are class-specific weights, $l_{ce}$ the cross-entropy loss, and $\gamma$ the focusing parameter adjusting the emphasis on hard examples
We used the AdamW optimizer \cite{loshchilov2017decoupled} ($\beta_1 = 0.9$, $\beta_2 = 0.999$), complemented by a linear learning rate warmup \cite{goyal2017accurate} for 1000 iterations, followed by a step-wise cosine learning rate decay schedule \cite{loshchilov2016sgdr}. The initial learning rate, after warmup, was $1\mathrm{e}{-5}$. Our training followed the configurations outlined in Sec.~\ref{sec:training_procedure}, adopting batch size-number of classes-number of shots configurations such as $(4, 1, 4)$, $(2, 4, 2)$, $(8, 1, 2)$, $(4, 2, 2)$, $(4, 4, 1)$, and $(16, 1, 1)$, where the batch size refers to a single GPU. The model, comprising 1.4 million learnable parameters (excluding the frozen backbone), underwent training for 50 epochs, with the iteration count per epoch equivalent to the size of $\mathcal{D}_{\text{train}}$. A cap was placed on point samples at a maximum of $10$ per instance to maintain efficiency.
Training leveraged the Leonardo cluster's resources, including 512GB of RAM and four NVIDIA A100-64GB GPUs.

\subsection{Results}

\begin{table}[t]
\centering
\caption{Results for $1$-way $K$-shot FSS segmentation, benchmarking against leading multi-class FSS methodologies and FPTrans.}
\small
  \begin{tabular}{c|c|c}
    \hline
    Method                          & $1$-way $1$-shot & $1$-way $5$-shot \\ \hline
    PANet~\cite{wang2019panet}      & 23.0             & 33.8             \\
    PPNet(w/o U)~\cite{liu2020part} & 25.7             & 36.2             \\
    PPNet~\cite{liu2020part}        & 27.2             & 36.7             \\
    MFNet~\cite{zhang2022mfnet}     & 34.9             & 39.2             \\
    FPTrans~\cite{zhang2022feature}     & \textbf{42.0}     & \textbf{53.8}    \\
    LA                              & 39.7             & 41.2             \\           \hline
    \end{tabular}
\label{tab:1-K}
\end{table}

In Table~\ref{tab:1-K}, we report the performance of LA on the COCO-$20^i$ benchmark for binary classification, comparing it against other multi-way approaches. We also include FPTrans \cite{zhang2022feature}, a dedicated binary classification model, as a reference. LA achieves the best results among the multi-way methods and performs comparably to FPTrans in the 1-way 1-shot setting, indicating that the architectural extensions introduced in LA preserve their effectiveness even in binary scenarios. However, LA shows limitations in mIoU when using multiple shots for the target class, underperforming compared to FPTrans. This highlights a trade-off: LA's flexibility comes at the cost of reduced specialization compared to methods tailored for 1-way settings. Nevertheless, like other prototype-based methods, LA underperforms compared to affinity-based approaches such as DCAMA~\cite{shi2022dense}, which leverages multi-scale features and performs dense pixel-wise comparisons---achieving, for example, a score of 50.9 in the 1-way 1-shot setting. However, affinity-based methods face scalability challenges in $N$-way $K$-shot scenarios, as the number of comparisons grows quadratically with the number of pixels. Specifically, if $N_q$ and $N_s$ denote the number of pixels in the query and support images, the number of pairwise comparisons is $N_q \times N_s$. This computational overhead highlights the practical advantages of prototype-based models, such as LA, for segmentation tasks involving multiple classes and shots.

\begin{figure}[tb]
    \centering
    \includegraphics[width=\linewidth]{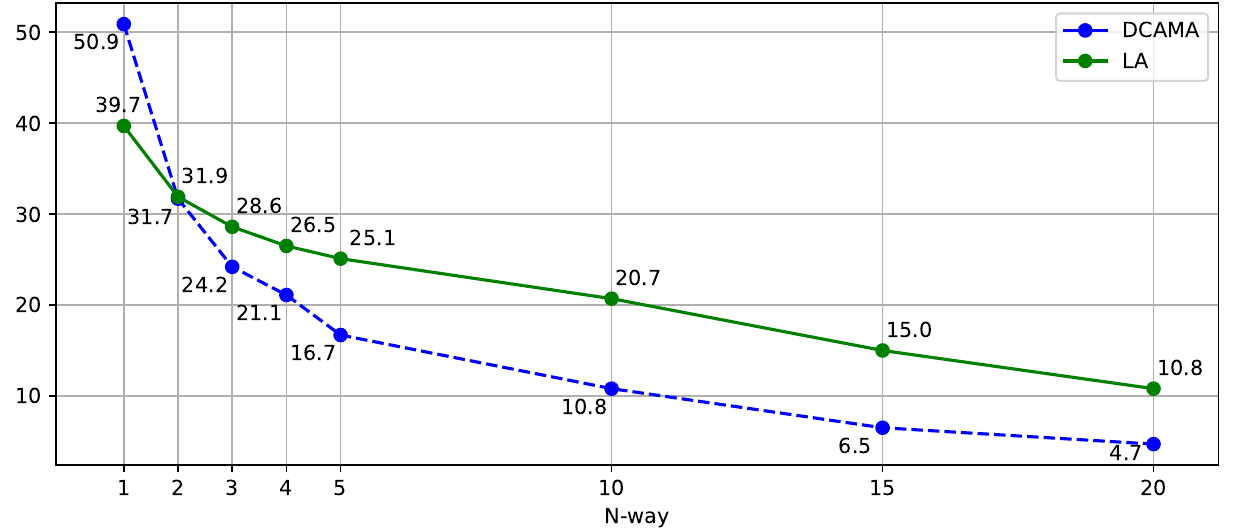}
    \caption{Average mIoU results for $N$-way 1-shot on COCO-$20^i$.}\label{fig:high_n}
\end{figure}

We tested LA’s performance under more complex conditions in increasingly larger $N$-way settings to further evaluate it. We also adapted DCAMA to a multi-class sequential variant to mitigate its original quadratic complexity and to assess whether it can maintain performance in this more scalable setup. Figure~\ref{fig:high_n} presents the results of LA and the adapted DCAMA on the $N$-way 1-shot segmentation task. Except for the 1-way case, LA consistently outperforms DCAMA across all other settings, underscoring its robustness and adaptability as the number of classes increases. To our knowledge, this is the first time an FSS method has been scaled beyond five classes in the multi-class setting. This highlights a current limitation in the field and a promising direction for future investigation into the scalability and generalization of few-shot segmentation methods.

\begin{table}[tb]
\centering
\caption{Profiling comparison between DCAMA and Label Anything in a 5-way setting (GB/s).}
\resizebox{\linewidth}{!}{
\begin{tabular}{lcccc}
\hline
Model & 1-shot & 2-shot & 5-shot & 10-shot \\
\hline
DCAMA \cite{zhang2022feature} & 6.72 / 0.22 & 12.19 / 0.30 & 28.00 / 0.50 & 55.00 / 1.02 \\
LA    & 0.57 / 0.08 & 00.68 / 0.10 & 00.99 / 0.14 & 01.50 / 0.21 \\
\hline
\end{tabular}
}
\label{tab:profiling_comparison}
\end{table}

\begin{table*}[t]
\centering
\small
\caption{
Performance on FSS benchmarks in $2$-way/$5$-way $1$-shot setting with multi-class FSS methods as re-evaluated in the MFNet study \cite{zhang2022mfnet}. Comparisons are made using mIoU percentages, specifically focusing on foreground classes. Misclassifications within these classes are considered under false positive and false negative categories for the respective classes, a measurement approach denoted as mIoU* in the MFNet analysis.
}
\label{tab:n-way}
% \resizebox{0.5\textwidth}{!}{
\begin{tabular}{cccccc|ccccc}
\hline
\multirow{2}{*}{Method} &
  \multicolumn{5}{c|}{$2$-way $1$-shot} &
  \multicolumn{5}{c}{$5$-way $1$-shot} \\ \cline{2-11} 
 &
  \multicolumn{1}{c|}{fold-0} &
  \multicolumn{1}{c|}{fold-1} &
  \multicolumn{1}{c|}{fold-2} &
  \multicolumn{1}{c|}{fold-3} &
  mean &
  \multicolumn{1}{c|}{fold-0} &
  \multicolumn{1}{c|}{fold-1} &
  \multicolumn{1}{c|}{fold-2} &
  \multicolumn{1}{c|}{fold-3} &
  mean \\ \hline
\multicolumn{1}{c|}{PANet \cite{wang2019panet}} &
  \multicolumn{1}{c|}{25.7} &
  \multicolumn{1}{c|}{16.1} &
  \multicolumn{1}{c|}{16.2} &
  \multicolumn{1}{c|}{13.8} &
  18.0 &
  \multicolumn{1}{c|}{22.1} &
  \multicolumn{1}{c|}{16.0} &
  \multicolumn{1}{c|}{14.1} &
  \multicolumn{1}{c|}{11.9} &
  16.0 \\
\multicolumn{1}{c|}{PPNet(w/o U)~\cite{liu2020part}} &
  \multicolumn{1}{c|}{29.0} &
  \multicolumn{1}{c|}{19.4} &
  \multicolumn{1}{c|}{16.5} &
  \multicolumn{1}{c|}{14.2} &
  19.8 &
  \multicolumn{1}{c|}{24.3} &
  \multicolumn{1}{c|}{16.8} &
  \multicolumn{1}{c|}{14.3} &
  \multicolumn{1}{c|}{12.8} &
  17.1 \\
\multicolumn{1}{c|}{PPNet~\cite{liu2020part}} &
  \multicolumn{1}{c|}{29.8} &
  \multicolumn{1}{c|}{19.7} &
  \multicolumn{1}{c|}{17.0} &
  \multicolumn{1}{c|}{15.1} &
  20.4 &
  \multicolumn{1}{c|}{25.6} &
  \multicolumn{1}{c|}{17.3} &
  \multicolumn{1}{c|}{15.5} &
  \multicolumn{1}{c|}{13.4} &
  18.0 \\
\multicolumn{1}{c|}{DENet~\cite{liu2020dynamic}} &
  \multicolumn{1}{c|}{30.3} &
  \multicolumn{1}{c|}{20.5} &
  \multicolumn{1}{c|}{16.7} &
  \multicolumn{1}{c|}{15.3} &
  20.7 &
  \multicolumn{1}{c|}{27.0} &
  \multicolumn{1}{c|}{17.9} &
  \multicolumn{1}{c|}{14.6} &
  \multicolumn{1}{c|}{14.0} &
  18.4 \\
\multicolumn{1}{c|}{MFNet~\cite{zhang2022mfnet}} &
  \multicolumn{1}{c|}{\textbf{35.3}} &
  \multicolumn{1}{c|}{24.0} &
  \multicolumn{1}{c|}{18.4} &
  \multicolumn{1}{c|}{18.7} &
  24.1 &
  \multicolumn{1}{c|}{\textbf{29.7}} &
  \multicolumn{1}{c|}{21.2} &
  \multicolumn{1}{c|}{15.4} &
  \multicolumn{1}{c|}{17.1} &
  20.9 \\
\multicolumn{1}{c|}{LA} &
  \multicolumn{1}{c|}{30.6} &
  \multicolumn{1}{c|}{\textbf{33.6}} &
  \multicolumn{1}{c|}{\textbf{32.2}} &
  \multicolumn{1}{c|}{\textbf{31.2}} &
  \textbf{31.9} &
  \multicolumn{1}{c|}{21.1} &
  \multicolumn{1}{c|}{\textbf{25.1}} &
  \multicolumn{1}{c|}{\textbf{26.4}} &
  \multicolumn{1}{c|}{\textbf{27.5}} &
  \textbf{25.1}\\ \hline
\end{tabular}%
% }
\end{table*}

The efficiency advantage of LA is further highlighted when increasing the number of shots in a 5-way setting. As shown in Table~\ref{tab:profiling_comparison}, DCAMA exhibits a near-linear growth in memory usage, reaching 55\,GB at 10 shots. In contrast, LA demonstrates a sublinear trend, requiring only 1.5\,GB at the same shot count.

Table \ref{tab:n-way} presents the performance of LA and several other state-of-the-art methods on the $2$-way $K$-shot segmentation task. LA demonstrates superior performance across all four folds, excelling in the more challenging $2$-way $1$-shot and $2$-way $5$-shot scenarios, achieving the highest mIoU scores. Notably, LA surpasses MFNet \cite{zhang2022mfnet}, previously regarded as the leading method, especially in folds 1, 2, and 3, underscoring its superior ability to handle multi-class segmentation tasks with limited examples.

We further evaluated LA in a non-exclusive class setting, where training and testing classes are not disjoint, as explored in \cite{wang2023seggpt} (Generalist Models). For a fair comparison, LA was trained using features extracted from ViT-L, consistent with the backbone used by Painter \cite{wangImagesSpeakImages2023} and SegGPT \cite{wang2023seggpt}. In this setting, LA demonstrates competitive performance, achieving state-of-the-art results in the 1-shot scenario and maintaining strong performance in the 5-shot scenario.

\begin{figure*}[tb]
    \centering
    \includegraphics[width=\linewidth]{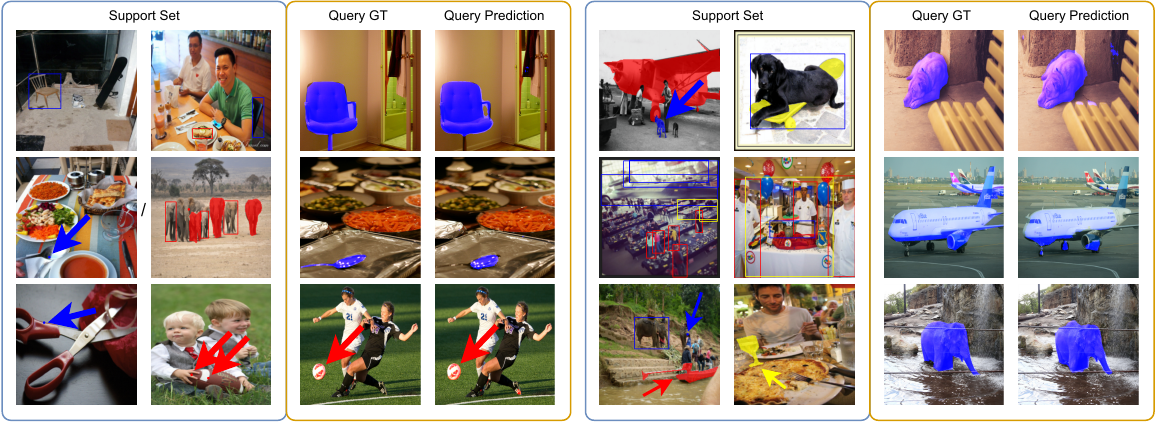}
  \caption{(Left) Visual representation of $2$-way $1$-shot segmentation on COCO-$20^i$. (Right) Visual representation of $2$-way, $L=3$ segmentation on COCO-$20^i$. The sequence from left to right includes the first prompt, the second prompt, the ground truth mask, and our model's prediction.}\label{fig:qualitative-results}
\end{figure*}

Figure~\ref{fig:qualitative-results} (left) shows qualitative results for the standard $N=2$, $K=1$ segmentation setting on the COCO-$20^i$ dataset. On the right, we present segmentations for a more complex scenario with $N=3$ and $L=2$, where prompts for three classes are distributed across two support examples. This setup allows greater expressiveness and compositional freedom in defining segmentation tasks. Furthermore, this visual representation highlights the efficacy of our model in accurately interpreting and responding to diverse prompts, showcasing its robust segmentation capabilities in complex scenarios.

\begin{table}[tb]
\centering
\caption{Quantitative results on COCO-$20^i$ obtained by having categories in training covering categories in testing.}
\small
\begin{tabular}{lcc}
\hline
Method  & 1-way 1-shot & 1-way 5-shot \\ \hline
\multicolumn{3}{c}{\textit{Specialist Models}}                             \\ \hline
HSNet            & 41.7                       & 50.7                       \\
VAT              & 42.9                       & 49.4                       \\
FPTrans          & 56.5                       & 65.5                       \\
LA & 50.5                       & 53.5                       \\
LA (L) & \textbf{57.4}              & 60.2                       \\ \hline
\multicolumn{3}{c}{\textit{Generalist Models}}                             \\ \hline
Painter          & 32.8                       & 32.6                       \\
SegGPT           & 56.1                       & \textbf{67.9}                       \\ \hline
\end{tabular}
\label{tab:train_results}
\end{table}

\subsection{Ablation Study}

Our comprehensive ablation studies on the validation fold of COCO-$20^i$ examined the contribution of each component within our model to its overall performance. We systematically omitted non-essential elements---specifically, the spatial convolutions in the mask decoder, the class token pool, and the class-example mixer from the prototype learner---and observed the resultant impact on model efficacy. As delineated in Table \ref{tab:ablation}, excluding the spatial convolutions from the mask decoder markedly diminished model performance. This component is pivotal for blending spatial details before classification, significantly enhancing segmentation quality. Meanwhile, the class token pool and class-example mixer, though their removal slightly affected performance, were proven indispensable for optimizing results. These findings highlight the integral role of each component in achieving superior segmentation outcomes, underscoring the holistic synergy that drives the performance of our model.
Our analysis also assessed the impact of different prompt types on the effectiveness of our model, exploring how each type influences its performance. Table \ref{tab:prompt_types} reveals that mask prompts led to the highest performance, offering the most detailed guidance for segmentation tasks. Nonetheless, the model demonstrated robust adaptability, yielding solid results even with alternative prompt types, albeit with a marginal reduction compared to masks alone. This underscores the model's versatile capability to manage and interpret various prompt types effectively.

\section{Conclusion}
\label{sec:conclusion}

In this study, we presented Label Anything, an innovative framework for few-shot semantic segmentation that advances the adaptability and efficiency of segmentation models. Our rigorous testing on the COCO-$20^i$ benchmark has demonstrated the model's capacity to achieve competitive and, in some instances, state-of-the-art performance, while also showcasing its ability to generalize across diverse segmentation scenarios. Including a wide range of visual prompts---points, bounding boxes, and masks---empowers Label Anything to effectively interpret and leverage multiple forms of input guidance. The results highlight the efficacy of prototype-based approaches in addressing the challenges of multi-class FSS, presenting a compelling alternative to affinity-based methods, which, despite their accuracy, struggle with scalability as the dataset expands. 

We envision several future research directions to develop our method further. An intriguing avenue involves a prototype-affinity multi-class hybrid method to achieve both efficiency and accuracy. 
This approach, coupled with the development of algorithms for auxiliary image search through similarity-based methods, aims to augment the support set with highly relevant images, thereby enhancing the model's adaptability to diverse scenarios. 
% Concurrently, the proposal of an optimized example candidate selection mechanism, potentially through clustering approaches, promises to streamline the identification of example candidates within large-scale datasets, enhancing the efficiency of support set construction. 
Concurrently, the proposal of an optimized example candidate selection mechanism promises to streamline candidate identification within large-scale datasets.

\begin{table}[tb]
    \centering
    \small
    \caption{Ablation analysis on the COCO-$20^i$'s fold-0 evaluating the impact of different components of the LA architecture.}
    \begin{tabular}{@{}c|c|c@{}}
    \hline
    Configuration                                     & 1-way 1-shot & 2-way 1-shot \\ \hline
    \multicolumn{1}{c|}{Complete model}           & 39.2 & 30.9 \\ 
    \multicolumn{1}{c|}{w/o Spatial convolutions} & 34.2 & 27.6 \\ 
    \multicolumn{1}{c|}{w/o Class-example mixer}  & 38.7 & 30.9 \\ 
    \multicolumn{1}{c|}{w/o Token pool}           & 38.3 & 30.1 \\ \hline
    \end{tabular}
\label{tab:ablation}
\end{table}

\begin{table}[t]
    \centering
    \small
    \caption{Ablation analysis on the impact of different prompt types on model performance.}\label{tab:prompt_types}
    \resizebox{\linewidth}{!}{
    \begin{tabular}{@{}c|c|c|c|c@{}}
    \hline
    Prompt Type      & 1-way 1-shot & 1-way 5-shot & 2-way 1-shot & 2-way 5-shot \\ \hline
    Only masks       & 39.7  & 41.2  & 31.9  & 32.5  \\
    Boxes and points & 37.1  & 40.5  & 29.2  & 30.7  \\
    Only boxes       & 38.3  & 40.9  & 30.6  & 31.5  \\
    Only points      & 36.5  & 39.9  & 29.0  & 29.4  \\ \hline
    \end{tabular}
    }
\end{table}

\newpage

\begin{ack}
We acknowledge ISCRA for awarding this project access to the LEONARDO supercomputer, owned by the EuroHPC Joint Undertaking, hosted by CINECA (Italy).
\end{ack}

%%%%%%%%%%%%%%%%%%%%%%%%%%%%%%%%%%%%%%%%%%%%%%%%%%%%%%%%%%%%%%%%%%%%%%%%

%%% Use this command to include your bibliography file.

\bibliography{main}

\end{document}